\documentclass[10pt,conference]{IEEEtran}

\usepackage{eso-pic}
\usepackage{tikz}

\newcommand{\copyrighttext}{\copyright~Ivan Petrukha, Yana Kurliak, Nataliia Stulova, accepted for publication by IEEE.
This version is provided under a Creative Commons Attribution-NonCommercial-NoDerivatives 4.0 International License (CC BY-NC-ND 4.0).}

\AddToShipoutPictureBG*{%
    \AtPageUpperLeft{%
        \setlength{\unitlength}{1in}%
        \begin{picture}(0,0)%
            \put(0.85,-0.55){%
                \begin{tikzpicture}
                    \node[draw, rectangle, thick, text width=\dimexpr\textwidth-0.2in\relax, align=center]{\copyrighttext};
                \end{tikzpicture}
            }%
        \end{picture}%
    }%
}

\usepackage{graphicx} 
\usepackage{amsmath,amssymb,amsfonts}
\usepackage{listings}
\usepackage{multirow}
\usepackage{booktabs}
\usepackage{siunitx}
\usepackage{xcolor}
\usepackage{float}
\usepackage{xspace}
\usepackage{textcomp}
\usepackage{cite}
\usepackage{tcolorbox}
\usepackage{amsmath}
\usepackage{orcidlink}

\usepackage{hyperref}
\hypersetup{hidelinks}

\def\BibTeX{{\rm B\kern-.05em{\sc i\kern-.025em b}\kern-.08em
    T\kern-.1667em\lower.7ex\hbox{E}\kern-.125emX}}

\definecolor{keywordcolor}{rgb}{0.0, 0.0, 1.0}
\definecolor{commentcolor}{rgb}{0.0, 0.5, 0.0}
\definecolor{light-gray}{gray}{0.95}

\newcommand{\code}[1]{\colorbox{light-gray}{\texttt{#1}}}
\newcommand{\graybox}[1]{\colorbox{gray!5}{#1}}
\newcommand{\greenbox}[1]{\colorbox{green!40}{#1}}
\newcommand{\lightgreenbox}[1]{\colorbox{green!15}{#1}}
\newcommand{\yellowbox}[1]{\colorbox{yellow!40}{#1}}
\newcommand{\lightyellowbox}[1]{\colorbox{yellow!15}{#1}}
\newcommand{\redbox}[1]{\colorbox{red!40}{#1}}

\newenvironment{tablefont}{\fontsize{9pt}{14pt}\selectfont}{}

\newcommand{\etal}{\textit{et~al.}\xspace}

\newcommand{\passat}[1]{\emph{pass@#1}} 

\begin{document}


\title{SwiftEval: Developing a Language-Specific Benchmark for LLM-generated Code Evaluation}


\author{
\IEEEauthorblockN{Ivan Petrukha~\orcidlink{0009-0009-3427-6869}}
\IEEEauthorblockA{\textit{MacPaw} \\
Kyiv, Ukraine \\
\textsf{\small petrukha@macpaw.com}}
\and
\IEEEauthorblockN{Yana Kurliak~\orcidlink{0009-0006-1073-0746}}
\IEEEauthorblockA{\textit{MacPaw} \\
Kyiv, Ukraine \\
\textsf{\small kurlyana@macpaw.com}}
\and
\IEEEauthorblockN{Nataliia Stulova~\orcidlink{0000-0002-6804-2253}}
\IEEEauthorblockA{\textit{MacPaw} \\
Kyiv, Ukraine \\
\textsf{\small nata.stulova@macpaw.com}}
}


\maketitle


\begin{abstract}
    In recent years, large language models (LLMs) have showcased significant advancements in code generation. 
    However, most evaluation benchmarks are primarily oriented towards Python, making it difficult to evaluate other programming languages, such as Swift, with high quality. 
    By examining widely established multilingual benchmarks like \textit{HumanEval-XL} and \textit{MultiPL-E}, we identified critical issues specific to their Swift components, making them insufficient or even irrelevant for assessing LLM coding capabilities on Swift.
    Unlike these existing approaches, which prioritize rapid scaling and generalization by automatically translating Python-centric benchmarks with LLMs, we adopt a quality-over-quantity methodology.
    We present SwiftEval, the first Swift-oriented benchmark consisting of 28 carefully hand-crafted problems, and evaluate 44 popular Code LLMs on it. 
    Our results show significant LLM scores drop for problems requiring language-specific features, most noticeable in the models of smaller sizes.
\end{abstract}




\section{Introduction}
\label{sec:intro}

Over the past few years, Large Language Models for Code (Code LLMs) \cite{2021-HumanEval-PassAtK-Codex,nijkamp_codegen_2022,xu_systematic_2022,guo_deepseek-coder_2024,roziere_code_2024} have significantly impacted the programming and software engineering fields, excelling in tasks such as code completion~\cite{wang_code_2021}, translation~\cite{pan_lost_2024}, summarization~\cite{ahmed_few-shot_2022}, and more. 
The advancement of Code LLMs in the natural language-to-code translation task~\cite{2021-HumanEval-PassAtK-Codex,2021-MBPP-and-MathQA,athiwaratkun_multi-lingual_2023,zan_large_2023} is particularly notable, reflecting the growing attention to this area.
To drive progress in the field of neural code generation, it is crucial to develop robust and high-quality evaluation methods.
Recent approaches have primarily focused on evaluating English-to-Python code generation (e.g. \cite{2021-HumanEval-PassAtK-Codex,2021-MBPP-and-MathQA,hendrycks_measuring_2021,nijkamp_codegen_2023}).
While there has been a growing interest in expanding evaluation to other programming~\cite{cassano_multipl-e_2022,athiwaratkun_multi-lingual_2023} and natural~\cite{peng_humaneval-xl_2024} languages, typically established multilingual benchmarks
incorporate other programming languages by leveraging LLMs capabilities to translate human-written Python problems.
However, such translation pipelines often lack careful consideration of the unique features and characteristics of other programming languages,
such as 
compiled vs. interpreted languages, static vs. dynamic typing, object-oriented vs. functional programming paradigms, and low-level vs. high-level languages, to name a few. 
To ensure the high-quality development of multilingual Code LLMs, evaluations must be tailored to each programming language, respecting its unique features.

To demonstrate the effect of such tailoring, we are taking 
the Swift programming language, and analyze Code LLMs on established multilingual benchmarks for it.
Swift is a statically typed, compiled programming language introduced by Apple in 2014 and getting steady adoption among Apple developers, in 2024 being ranked in top-20 in the TIOBE~\cite{noauthor_tiobe_nodate} index and residing in the third quartile of the RedMonk ranking~\cite{ogrady_redmonk_2024}.
We analyzed state-of-the-art multilingual benchmarks covering the Swift programming language, including 
\textit{MBXP}~\cite{athiwaratkun_multi-lingual_2023}, 
\textit{HumanEval-XL}~\cite{peng_humaneval-xl_2024}, 
and 
\textit{MultiPL-E}~\cite{cassano_multipl-e_2022}.
Our experiments demonstrate that these benchmarks fall short in providing a comprehensive and high-quality evaluation of Swift code generation due to their Python-centric design.
We provide a detailed discussion of the issues encountered -- issues that extend to other programming languages --- and include practical examples.
To address these limitations, we present SwiftEval, the first Swift-oriented benchmark consisting of 28 carefully hand-crafted problems, and evaluate 44 popular Code LLMs on it.
We report the comparative results of the evaluations conducted on both the established HumanEval benchmark and our SwiftEval.
Our findings demonstrate that even a small, but language-tailored benchmark can provide more insightful results than large, popular, but general-purpose benchmarks, advancing the programming language understanding capabilities of large language models.


\section{Background and Motivation}
\label{sec:background}

OpenAI's \emph{HumanEval}~\cite{2021-HumanEval-PassAtK-Codex} benchmark is widely used to measure the functional correctness of generated code.
HumanEval consists of 164 programming problems, assessing Python language comprehension, algorithms, and mathematics. 
The problems include a function signature, docstring, body, and several unit tests. 
Instead of relying on standard machine translation metrics like BLEU~\cite{papineni_bleu_2001} or CodeBLEU~\cite{ren_codebleu_2020} to compare the ground truth code and the LLM-generated one, authors introduce the \textit{pass@k} metric to calculate scores based on a number of unit tests that the generated code passes.
However, HumanEval popularity has raised concerns about potential memorization:
Yang~\etal~\cite{yang_rethinking_2023} report that pre-training sets such as RedPajama-Data-1T and StarCoder-Data overlap with HumanEval by 8--18\%.
This issue can compromise the benchmark-based evaluation due to model overfitting, an effect we observe in \textit{Section}~\ref{sec:evaluation}.
Following HumanEval, the Google Research team proposed the \emph{Most Basic Python Problems} (MBPP)~\cite{2021-MBPP-and-MathQA}, another widely used benchmark. 
It consists of 974 programming problems, designed to test LLM synthesis capabilities from English to Python with focus on programming fundamentals and standard library functionality.
In recent years, many fresh benchmarks containing problem and test sets and relying on the \passat{k} metric 
were proposed for data science tasks~\cite{lai_ds-1000_2022}, class-level code generation~\cite{du_classeval_2023}, cross-file code completion~\cite{ding_crosscodeeval_2023}, and more.

Gradual extension of benchmarks to other programming languages had its issues due to direct translations from Python to another language with unique features and syntax that don't align with Python.
For instance, MBXP~\cite{athiwaratkun_multi-lingual_2023} classifies languages into 'in-domain' (present in training datasets, such as Python, Java, and JavaScript) and 'out-of-domain' (e.g., C++, Ruby, Swift), as primary motivation of the research was to assess the generalization ability of language models on out-of-domain languages.
While in-domain languages receive careful translations and manual validation, out-of-domain languages often lack such attention, resulting in benchmarks that contain numerous errors. Moreover, even among in-domain languages, challenges persist.
For example, \cite{cao_javabench_2024} emphasizes Java's object-oriented features and underscores the importance of evaluating advanced programming concepts. 
It critiques the relevance of directly translating Python-oriented benchmarks, which prioritize functional programming and basic coding skills, to an object-oriented language like Java.

For Swift the situation is even worse.
Only 4 of 27 discovered benchmarks~\cite{cao_javabench_2024,raihan_mhumaneval_2024} support Swift, although only as an additional multilingual translation.
On top of that, problems not only miss commonly used language features, but include critical translation mistakes that make the assessment invalid.
For instance, HumanEval-XL~\cite{peng_humaneval-xl_2024} is a multilingual code generation benchmark that establishes connections between 23 natural languages and 12 programming languages. 
Although it includes Swift, a closer look reveals significant translation issues.
For instance, 27\% of problems use \code{AnyHashable} as input argument type, which breaks Swift's type system and confuses language models.
Additionally, the translations ignore essential Swift features like optional values, forcing models to generate workarounds and return \code{"none"} instead of conventional \code{nil}.
Due to these issues, we found HumanEval-XL unsuitable for Swift code quality assessment.
Another benchmark, MultiPL-E~\cite{cassano_multipl-e_2022}, translates HumanEval and MBPP into 18 other programming languages, including Swift.
The most popular multilingual leaderboard \emph{"Big Code Models Leaderboard"}~\footnote{\url{https://huggingface.co/spaces/bigcode/bigcode-models-leaderboard}} is based on this benchmark.
Despite its popularity, we found that 8 problems from the HumanEval subset weren't solved even once.
These failures were due to various reasons, such as the 
\code{operator} reserved keyword usage as function argument label or misaligned unit tests.
For instance, one task expects the solution to round the answer to the nearest integer using \code{round} function, but its test cases assume Python’s bankers rounding, which rounds to the nearest even integer. 
For example, rounding 8.5 yields 8 in Python but 9 in C++, JavaScript, or Swift.
This highlights the Python-centric design of HumanEval and its unsuitability for other languages.
In the MultiPL-E (MBPP) subset, we also found a lot of critical issues (10\% of problems) with tuples.
As an example, we want to highlight one task where an array of integers should be expanded into a tuple.
This task is completely unsolvable in Swift because tuples in Swift have fixed sizes, unlike Python's dynamic ones.

\begin{table*}[htp!]
\centering
\caption{
\vspace{6pt}
Code Generation Evaluation Comparison of HumanEval (Python) and SwiftEval (Swift) Benchmarks. 
Color coding indicates performance changes with \graybox{same level}, \lightgreenbox{minor gain}, \greenbox{major gain}, \lightyellowbox{minor drop}, \yellowbox{major drop}, \redbox{critical drop}.
}
\setlength{\tabcolsep}{10pt}
\renewcommand{\arraystretch}{1.0}
\begin{tablefont}
\begin{tabular}{||l|l|c c|c c c|c c||}
\hline 
\multicolumn{2}{||c|}{Experiment} &
\multicolumn{2}{c|}{HumanEval} &
\multicolumn{3}{c|}{SwiftEval} &
\multicolumn{2}{c||}{Difference} \\
\hline
Model & Size & Score & Rank & Score & Variance & Rank & Score & Rank \\
\hline
GPT-4o &- &90.3 &2 &88.9 &±2.6 &1 &\graybox{-1.4} &\graybox{1} \\
GPT-4o Mini &- &87.2 &7 &85.6 &±2.9 &3 &\graybox{-1.6} &\lightgreenbox{4} \\
GPT-4 Turbo &- &88.2 &6 &87.1 &±2.7 &2 &\graybox{-1.1} &\lightgreenbox{4} \\
GPT-4 &- &86.6 &8 &82.2 &±3.2 &5 &\lightyellowbox{-4.4} &\lightgreenbox{3} \\
GPT-3.5 Turbo &- &68.0 &21 &81.3 &±3.2 &6 &\greenbox{13.3} &\greenbox{15} \\
\hline
Codestral Mamba &7B &75.0 &19 &58.9 &±4.1 &11 &\yellowbox{-16.1} &\greenbox{8} \\
Codestral &22B &81.1 &13 &77.8 &±3.4 &8 &\graybox{-3.3} &\lightgreenbox{5} \\
\hline
CodeLlama Instruct &7B &34.8 &41 &28.5 &±3.8 &27 &\lightyellowbox{-6.3} &\greenbox{14} \\
CodeLlama Instruct &13B &42.7 &37 &43.7 &±3.9 &16 &\graybox{1.0} &\greenbox{21} \\
CodeLlama Instruct &34B &41.5 &38 &42.1 &±3.9 &17 &\graybox{0.6} &\greenbox{21} \\
CodeLlama Instruct &70B &67.8 &22 &52.3 &±4.0 &14 &\yellowbox{-15.5} &\greenbox{8} \\
\hline
CodeGemma &2B &31.1 &44 &15.2 &±2.9 &37 &\yellowbox{-15.9} &\greenbox{7} \\
CodeGemma &7B &44.5 &35 &26.9 &±3.8 &30 &\yellowbox{-17.6} &\lightgreenbox{5} \\
CodeGemma Instruct &7B &56.1 &31 &28.1 &±3.7 &28 &\redbox{-28.0} &\lightgreenbox{3} \\
CodeGemma 1.1 Instruct &7B &60.4 &27 &36.2 &±4.1 &22 &\yellowbox{-24.2} &\lightgreenbox{5} \\
\hline
CodeGeeX2 &6B &35.9 &39 &2.5 &±1.3 &44 &\redbox{-33.4} &\lightyellowbox{-5} \\
CodeGeeX4 &9B &82.3 &12 &53.7 &±3.9 &12 &\redbox{-28.6} &\lightgreenbox{0} \\
\hline
CodeQwen1.5 &7B &51.8 &32 &41.5 &±3.8 &18 &\lightyellowbox{-10.3} &\greenbox{14} \\
CodeQwen1.5 Chat &7B &83.5 &11 &37.9 &±3.9 &21 &\redbox{-45.6} &\yellowbox{-10} \\
\hline
Qwen2.5 Coder &1.5B &43.9 &36 &11.3 &±2.7 &39 &\redbox{-32.6} &\lightyellowbox{-3} \\
Qwen2.5 Coder &7B &61.6 &26 &25.5 &±3.5 &31 &\redbox{-36.1} &\lightyellowbox{-5} \\
Qwen2.5 Coder Instruct &7B &88.4 &5 &40.0 &±3.8 &19 &\redbox{-48.4} &\redbox{-14} \\
Qwen2.5 Coder Instruct &14B &89.6 &4 &62.8 &±4.0 &10 &\redbox{-26.8} &\yellowbox{-6} \\
Qwen2.5 Coder Instruct &32B &92.7 &1 &79.1 &±3.4 &7 &\lightyellowbox{-13.6} &\yellowbox{-6} \\
\hline
DeepSeek Coder Instruct &1.3B &65.2 &24 &10.8 &±2.6 &40 &\redbox{-54.4} &\redbox{-16} \\
DeepSeek Coder Instruct &6.7B &78.6 &17 &17.5 &±3.1 &36 &\redbox{-61.1} &\redbox{-19} \\
DeepSeek Coder Instruct &33B &79.3 &15 &32.0 &±3.8 &25 &\redbox{-47.3} &\yellowbox{-10} \\
\hline
DeepSeek Coder V2 Instruct &16B &81.1 &13 &69.1 &±3.7 &9 &\lightyellowbox{-12.0} &\lightgreenbox{4} \\
DeepSeek Coder V2 Instruct &236B &90.2 &3 &82.4 &±3.2 &4 &\lightyellowbox{-7.8} &\lightyellowbox{-1} \\
\hline
Granite Code Instruct &3B &51.2 &33 &12.1 &±2.7 &38 &\redbox{-39.1} &\lightyellowbox{-5} \\
Granite Code Instruct &8B &57.9 &30 &23.1 &±3.6 &32 &\redbox{-34.8} &\lightyellowbox{-2} \\
Granite Code Instruct &20B &60.4 &27 &17.6 &±3.1 &35 &\redbox{-42.8} &\yellowbox{-8} \\
Granite Code Instruct &34B &62.2 &25 &32.1 &±3.9 &24 &\redbox{-30.1} &\graybox{1} \\
\hline
StarCoder 2 &3B &31.7 &43 &21.3 &±3.4 &33 &\lightyellowbox{-10.4} &\greenbox{10} \\
StarCoder 2 &7B &35.4 &40 &28.9 &±3.6 &26 &\lightyellowbox{-6.5} &\greenbox{14} \\
StarCoder 2 &15B &46.3 &34 &45.5 &±4.1 &15 &\graybox{-0.8} &\greenbox{19} \\
StarCoder 2 Instruct &15B &72.6 &20 &53.6 &±4.0 &13 &\yellowbox{-19.0} &\greenbox{7} \\
\hline
Stable Code &3B &32.4 &42 &9.1 &±2.5 &42 &\yellowbox{-23.3} &\graybox{0} \\
Stable Code Instruct &3B &59.0 &29 &10.2 &±2.4 &41 &\redbox{-48.8} &\redbox{-12} \\
\hline
OpenCodeInterpreter &6.7B &76.2 &18 &18.9 &±3.1 &34 &\redbox{-57.3} &\redbox{-16} \\
OpenCodeInterpreter &33B &79.3 &15 &27.3 &±3.5 &29 &\redbox{-52.0} &\redbox{-14} \\
\hline
Yi-Coder Chat &1.5B &67.7 &23 &4.9 &±1.7 &43 &\redbox{-62.8} &\redbox{-20} \\
Yi-Coder Chat &9B &85.4 &10 &33.9 &±3.9 &23 &\redbox{-51.5} &\redbox{-13} \\
\hline
Nxcode-CQ &7B &86.6 &8 &39.7 &±4.0 &20 &\redbox{-46.9} &\redbox{-12} \\
\hline
\end{tabular}
\label{tab:model_scores}
\end{tablefont}
\end{table*}

\section{Solution}
\label{sec:solution}

To address the limitations of existing code generation benchmarks, we decided to create our own benchmark.
We designed \textbf{SwiftEval}, a benchmark featuring manually crafted problems, rather than relying on translations from other benchmarks.
The benchmark is designed specially for the Swift programming language and considers unique Swift features like \textit{static typing}, \textit{protocols}, \textit{generics}, \textit{enumerations}, \textit{closures}, etc.
These Swift features are not covered by any other benchmark, so we found it important to focus on them.
In its current state, SwiftEval contains 28 unique problems.
The problems were designed by an industry Software Engineer with deep Swift knowledge.
Each problem has a natural language query, an additional code context to consider, a code entrypoint where generation starts, and 3-5 unit tests to validate correctness.
While HumanEval has only function-level tasks, we combined both function-level and class-level approaches.
In addition, we focused on creating more practical problems, not only algorithmic ones.
Our benchmark covers a diverse set of problems, including tasks such as running system executables with arguments, calculating abstract file-system metadata, comparing semantic version strings, and implementing design patterns such as state machine, dependency injection container, cache with maximum capacity, among others.
Problem details, 
together with experiment results (\textit{Section}~\ref{sec:evaluation}) 
are available \emph{online}~\footnote{\url{https://doi.org/10.5281/zenodo.14445601}}.


\section{Evaluation}
\label{sec:evaluation}

We performed 44 experiments on open-source and closed-source models of different sizes.
For each experiment, we calculate a pass@1 score with temperature 0.2, token probability 0.95, and 20 completions for each problem.
While the OpenAI team used 200 completions for each problem to estimate pass@k reliably in~\cite{2021-HumanEval-PassAtK-Codex}, later work~\cite{cassano_multipl-e_2022} observed that pass@1 rates appear to stabilize around 20 samples, suggesting that future studies could achieve stable estimates with significantly lower computational overhead.
For open-source models, we used the default model prompt template provided in the tokenizer configuration when available. 
The use of the default model prompt template is important because it aligns with the model’s pretraining, ensuring optimal token handling.
We used the OpenAI API and HuggingFace Transformers library (version 4.45.2) for code generation. 
The generated code was compiled using the official Apple Swift compiler (version 5.10) and executed on a test machine running macOS 14.6.1.
%

Table \ref{tab:model_scores} shows a comparison of the evaluation results for the HumanEval and SwiftEval benchmarks.
Benchmark scores estimate a model’s true performance with variance due to finite sample sets, so we report it via 95\% confidence intervals calculated using the bootstrapping method.
A significant gap exists between OpenAI's ChatGPT models and open-source models.
Only 3 models came close to ChatGPT's performance: \textit{DeepSeek Coder V2 Instruct}~\cite{deepseek-ai_deepseek-coder-v2_2024} with 236B parameters, \textit{Qwen2.5 Coder Instruct}~\cite{hui_qwen25-coder_2024} with 32B parameters, and \textit{Codestral}~\cite{ai_codestral_2024} with 22B parameters.
Within our benchmark, small models did not perform well, and their scores dropped significantly, while bigger models stayed more stable.

\begin{figure}[h!]
    \centering
    \includegraphics[width=\columnwidth]{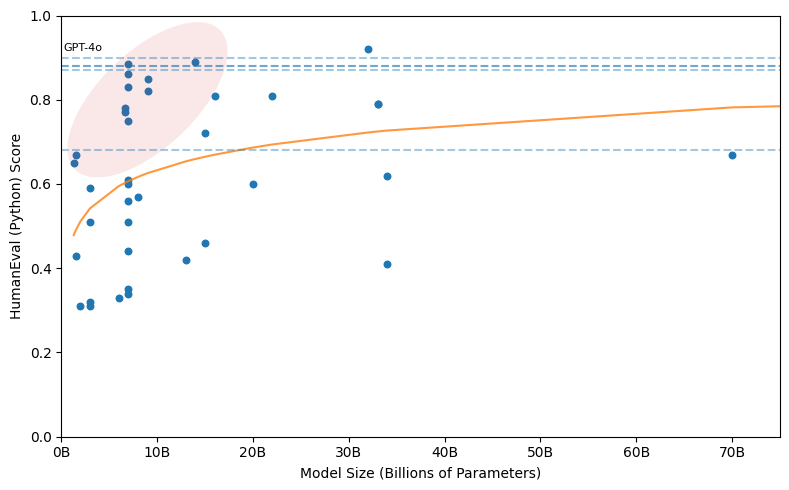}
    \vspace{-18pt}
    \caption{Performance on HumanEval (Python) Across Model Sizes}
    \label{fig:he_trendline_plot}
\end{figure}

Figure~\ref{fig:he_trendline_plot} with a dependency plot between the size of models we consider (under 70B parameters) and their evaluation scores for the HumanEval benchmark.
The blue points represent specific models, plotted according to their size and evaluation scores, while the blue lines denote models with unknown sizes.
The near-horizontal Ordinary Least Squares (OLS) trendline (orange) indicates a weak correlation between model size and performance,  where, unexpectedly, small models achieve scores comparable to state-of-the-art models such as OpenAI's ChatGPT (red ellipse area).

\begin{figure}[h!]
    \centering
    \includegraphics[width=\columnwidth]{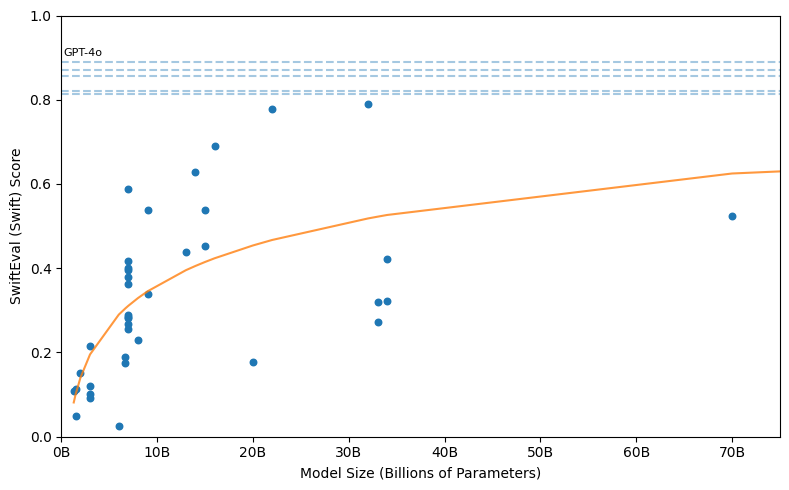}
    \vspace{-18pt}
    \caption{Performance on SwiftEval (Swift) Across Model Sizes}
    \label{fig:se_trendline_plot}
\end{figure}

Figure \ref{fig:se_trendline_plot} shows a plot with the same structure but within SwiftEval benchmark evaluations.
Here, we can see a more inclined trendline compared to the HumanEval version.
Such a trendline represents a better dependency between model size and score.
HumanEval shows a weak correlation (0.30) between model size and score, which supports the point of potential benchmark memorization during model training.
On the other hand, SwiftEval has a higher correlation (0.50) between model size and score, which makes us more confident in our results.
%


\section{Related Work}
\label{sec:related_work}

During the last year, many authors reported different drawbacks of the existing benchmarks.
In JavaBench \cite{cao_javabench_2024}, authors highlight that existing benchmarks primarily assess basic coding skills while overlooking advanced features.
In CPP-UT-Bench \cite{bhargava_cpp-ut-bench_2024}, authors report that the existing coding benchmarks mostly cover high-level languages, and they are far from real-world software engineering tasks.
In HumanEval.jl \cite{yi_team_humanevaljl_2024}, authors suggest improving MultiPL-E prompts translation because the original prompts are not that accurate and conventional regarding Julia programming language.
In Rust Compiling Benchmark \cite{gusanidas_rust_2024}, authors provide a benchmark for Rust programming language but only to check if the generated code compiles successfully. 
We find this method viable for compiled languages as it requires less time to design problems. 
In SwiftEval, 86\% of evaluations (12,518) failed due to compilation errors, while only 14\% (2,153) failed because of unit test failures.


\section{Limitations}
\label{sec:limitations}

We do not assess the security aspect of the generated code, so we strongly recommend running test executions in a virtual machine or another sandboxed environment for safety.
SwiftEval currently contains 28 problems, which is significantly fewer than HumanEval (164 problems) or MBPP (974 problems), so we assume that results may vary slightly when more problems are added.
This limitation results from the fact that it is much more difficult to create hand-crafted problems than to translate existing problems from other benchmarks.

\section{Conclusions and Future Work}
\label{sec:conclusions_and_future_work}

We find that the quality of popular multilingual benchmarks is generally inadequate with respect to the Swift programming language.
Therefore, we recommend that any future evaluation efforts that involve translated prompts carefully verify the quality of the translation for each specific programming language before reporting the results or drawing conclusions about a model's performance on the language of interest.
However, the bigger problem is the multilingual benchmark Python-centric base.
We agree with other authors about HumanEval limitations and suggest other authors create fresh micro benchmarks to test your specific programming language.
In future work we will address generated code secureness and efficiency.
We also plan to create subcategories (e.g. design patterns, system frameworks, user interfaces) to understand model strengths and weaknesses.
Such analysis will help to understand what current training datasets are missing.



\bibliographystyle{ieeetr}
\bibliography{zotero-references,webpage-links}

\begin{thebibliography}{10}

\bibitem{2021-HumanEval-PassAtK-Codex}
M.~Chen, J.~Tworek, H.~Jun, Q.~Yuan, H.~P. d.~O. Pinto, J.~Kaplan, H.~Edwards, Y.~Burda, N.~Joseph, G.~Brockman, A.~Ray, R.~Puri, G.~Krueger, M.~Petrov, H.~Khlaaf, G.~Sastry, P.~Mishkin, B.~Chan, S.~Gray, N.~Ryder, M.~Pavlov, A.~Power, L.~Kaiser, M.~Bavarian, C.~Winter, P.~Tillet, F.~P. Such, D.~Cummings, M.~Plappert, F.~Chantzis, E.~Barnes, A.~Herbert-Voss, W.~H. Guss, A.~Nichol, A.~Paino, N.~Tezak, J.~Tang, I.~Babuschkin, S.~Balaji, S.~Jain, W.~Saunders, C.~Hesse, A.~N. Carr, J.~Leike, J.~Achiam, V.~Misra, E.~Morikawa, A.~Radford, M.~Knight, M.~Brundage, M.~Murati, K.~Mayer, P.~Welinder, B.~McGrew, D.~Amodei, S.~McCandlish, I.~Sutskever, and W.~Zaremba, ``Evaluating {Large} {Language} {Models} {Trained} on {Code},'' July 2021.
\newblock arXiv:2107.03374 [cs].

\bibitem{nijkamp_codegen_2022}
E.~Nijkamp, B.~Pang, H.~Hayashi, L.~Tu, H.~Wang, Y.~Zhou, S.~Savarese, and C.~Xiong, ``{CodeGen}: {An} {Open} {Large} {Language} {Model} for {Code} with {Multi}-{Turn} {Program} {Synthesis},'' Sept. 2022.

\bibitem{xu_systematic_2022}
F.~F. Xu, U.~Alon, G.~Neubig, and V.~J. Hellendoorn, ``A {Systematic} {Evaluation} of {Large} {Language} {Models} of {Code},'' May 2022.
\newblock arXiv:2202.13169.

\bibitem{guo_deepseek-coder_2024}
D.~Guo, Q.~Zhu, D.~Yang, Z.~Xie, K.~Dong, W.~Zhang, G.~Chen, X.~Bi, Y.~Wu, Y.~K. Li, F.~Luo, Y.~Xiong, and W.~Liang, ``{DeepSeek}-{Coder}: {When} the {Large} {Language} {Model} {Meets} {Programming} -- {The} {Rise} of {Code} {Intelligence},'' Jan. 2024.
\newblock arXiv:2401.14196.

\bibitem{roziere_code_2024}
B.~Rozière, J.~Gehring, F.~Gloeckle, S.~Sootla, I.~Gat, X.~E. Tan, Y.~Adi, J.~Liu, R.~Sauvestre, T.~Remez, J.~Rapin, A.~Kozhevnikov, I.~Evtimov, J.~Bitton, M.~Bhatt, C.~C. Ferrer, A.~Grattafiori, W.~Xiong, A.~Défossez, J.~Copet, F.~Azhar, H.~Touvron, L.~Martin, N.~Usunier, T.~Scialom, and G.~Synnaeve, ``Code {Llama}: {Open} {Foundation} {Models} for {Code},'' Jan. 2024.
\newblock arXiv:2308.12950.

\bibitem{wang_code_2021}
Y.~Wang and H.~Li, ``Code {Completion} by {Modeling} {Flattened} {Abstract} {Syntax} {Trees} as {Graphs},'' {\em Proceedings of the AAAI Conference on Artificial Intelligence}, vol.~35, pp.~14015--14023, May 2021.

\bibitem{pan_lost_2024}
R.~Pan, A.~R. Ibrahimzada, R.~Krishna, D.~Sankar, L.~P. Wassi, M.~Merler, B.~Sobolev, R.~Pavuluri, S.~Sinha, and R.~Jabbarvand, ``Lost in {Translation}: {A} {Study} of {Bugs} {Introduced} by {Large} {Language} {Models} while {Translating} {Code},'' in {\em Proceedings of the {IEEE}/{ACM} 46th {International} {Conference} on {Software} {Engineering}}, (Lisbon Portugal), pp.~1--13, ACM, Apr. 2024.

\bibitem{ahmed_few-shot_2022}
T.~Ahmed and P.~Devanbu, ``Few-shot training {LLMs} for project-specific code-summarization,'' Sept. 2022.
\newblock arXiv:2207.04237.

\bibitem{2021-MBPP-and-MathQA}
J.~Austin, A.~Odena, M.~Nye, M.~Bosma, H.~Michalewski, D.~Dohan, E.~Jiang, C.~Cai, M.~Terry, Q.~Le, and C.~Sutton, ``Program {Synthesis} with {Large} {Language} {Models},'' Aug. 2021.
\newblock arXiv:2108.07732 [cs].

\bibitem{athiwaratkun_multi-lingual_2023}
B.~Athiwaratkun, S.~K. Gouda, Z.~Wang, X.~Li, Y.~Tian, M.~Tan, W.~U. Ahmad, S.~Wang, Q.~Sun, M.~Shang, S.~K. Gonugondla, H.~Ding, V.~Kumar, N.~Fulton, A.~Farahani, S.~Jain, R.~Giaquinto, H.~Qian, M.~K. Ramanathan, R.~Nallapati, B.~Ray, P.~Bhatia, S.~Sengupta, D.~Roth, and B.~Xiang, ``Multi-lingual {Evaluation} of {Code} {Generation} {Models},'' Mar. 2023.
\newblock arXiv:2210.14868.

\bibitem{zan_large_2023}
D.~Zan, B.~Chen, F.~Zhang, D.~Lu, B.~Wu, B.~Guan, Y.~Wang, and J.-G. Lou, ``Large {Language} {Models} {Meet} {NL2Code}: {A} {Survey},'' May 2023.
\newblock arXiv:2212.09420.

\bibitem{hendrycks_measuring_2021}
D.~Hendrycks, S.~Basart, S.~Kadavath, M.~Mazeika, A.~Arora, E.~Guo, C.~Burns, S.~Puranik, H.~He, D.~Song, and J.~Steinhardt, ``Measuring {Coding} {Challenge} {Competence} {With} {APPS},'' Nov. 2021.
\newblock arXiv:2105.09938.

\bibitem{nijkamp_codegen_2023}
E.~Nijkamp, B.~Pang, H.~Hayashi, L.~Tu, H.~Wang, Y.~Zhou, S.~Savarese, and C.~Xiong, ``{CodeGen}: {An} {Open} {Large} {Language} {Model} for {Code} with {Multi}-{Turn} {Program} {Synthesis},'' Feb. 2023.
\newblock arXiv:2203.13474.

\bibitem{cassano_multipl-e_2022}
F.~Cassano, J.~Gouwar, D.~Nguyen, S.~Nguyen, L.~Phipps-Costin, D.~Pinckney, M.-H. Yee, Y.~Zi, C.~J. Anderson, M.~Q. Feldman, A.~Guha, M.~Greenberg, and A.~Jangda, ``{MultiPL}-{E}: {A} {Scalable} and {Extensible} {Approach} to {Benchmarking} {Neural} {Code} {Generation},'' Dec. 2022.
\newblock arXiv:2208.08227 [cs].

\bibitem{peng_humaneval-xl_2024}
Q.~Peng, Y.~Chai, and X.~Li, ``{HumanEval}-{XL}: {A} {Multilingual} {Code} {Generation} {Benchmark} for {Cross}-lingual {Natural} {Language} {Generalization},'' Mar. 2024.
\newblock arXiv:2402.16694 [cs].

\bibitem{noauthor_tiobe_nodate}
``{TIOBE} {Index}.''

\bibitem{ogrady_redmonk_2024}
S.~O'Grady, ``The {RedMonk} {Programming} {Language} {Rankings}: {June} 2024,'' Sept. 2024.

\bibitem{papineni_bleu_2001}
K.~Papineni, S.~Roukos, T.~Ward, and W.-J. Zhu, ``{BLEU}: a method for automatic evaluation of machine translation,'' in {\em Proceedings of the 40th {Annual} {Meeting} on {Association} for {Computational} {Linguistics} - {ACL} '02}, (Philadelphia, Pennsylvania), p.~311, Association for Computational Linguistics, 2001.

\bibitem{ren_codebleu_2020}
S.~Ren, D.~Guo, S.~Lu, L.~Zhou, S.~Liu, D.~Tang, N.~Sundaresan, M.~Zhou, A.~Blanco, and S.~Ma, ``{CodeBLEU}: a {Method} for {Automatic} {Evaluation} of {Code} {Synthesis},'' Sept. 2020.
\newblock arXiv:2009.10297 [cs].

\bibitem{yang_rethinking_2023}
S.~Yang, W.-L. Chiang, L.~Zheng, J.~E. Gonzalez, and I.~Stoica, ``Rethinking {Benchmark} and {Contamination} for {Language} {Models} with {Rephrased} {Samples},'' Nov. 2023.
\newblock arXiv:2311.04850 [cs].

\bibitem{lai_ds-1000_2022}
Y.~Lai, C.~Li, Y.~Wang, T.~Zhang, R.~Zhong, L.~Zettlemoyer, S.~W.-t. Yih, D.~Fried, S.~Wang, and T.~Yu, ``{DS}-1000: {A} {Natural} and {Reliable} {Benchmark} for {Data} {Science} {Code} {Generation},'' Nov. 2022.
\newblock arXiv:2211.11501.

\bibitem{du_classeval_2023}
X.~Du, M.~Liu, K.~Wang, H.~Wang, J.~Liu, Y.~Chen, J.~Feng, C.~Sha, X.~Peng, and Y.~Lou, ``{ClassEval}: {A} {Manually}-{Crafted} {Benchmark} for {Evaluating} {LLMs} on {Class}-level {Code} {Generation},'' Aug. 2023.
\newblock arXiv:2308.01861.

\bibitem{ding_crosscodeeval_2023}
Y.~Ding, Z.~Wang, W.~U. Ahmad, H.~Ding, M.~Tan, N.~Jain, M.~K. Ramanathan, R.~Nallapati, P.~Bhatia, D.~Roth, and B.~Xiang, ``{CrossCodeEval}: {A} {Diverse} and {Multilingual} {Benchmark} for {Cross}-{File} {Code} {Completion},'' Nov. 2023.
\newblock arXiv:2310.11248.

\bibitem{cao_javabench_2024}
J.~Cao, Z.~Chen, J.~Wu, S.-c. Cheung, and C.~Xu, ``{JavaBench}: {A} {Benchmark} of {Object}-{Oriented} {Code} {Generation} for {Evaluating} {Large} {Language} {Models},'' Oct. 2024.
\newblock arXiv:2406.12902.

\bibitem{raihan_mhumaneval_2024}
N.~Raihan, A.~Anastasopoulos, and M.~Zampieri, ``{mHumanEval} -- {A} {Multilingual} {Benchmark} to {Evaluate} {Large} {Language} {Models} for {Code} {Generation},'' Oct. 2024.
\newblock arXiv:2410.15037.

\bibitem{deepseek-ai_deepseek-coder-v2_2024}
{DeepSeek-AI}, Q.~Zhu, D.~Guo, Z.~Shao, D.~Yang, P.~Wang, R.~Xu, Y.~Wu, Y.~Li, H.~Gao, S.~Ma, W.~Zeng, X.~Bi, Z.~Gu, H.~Xu, D.~Dai, K.~Dong, L.~Zhang, Y.~Piao, Z.~Gou, Z.~Xie, Z.~Hao, B.~Wang, J.~Song, D.~Chen, X.~Xie, K.~Guan, Y.~You, A.~Liu, Q.~Du, W.~Gao, X.~Lu, Q.~Chen, Y.~Wang, C.~Deng, J.~Li, C.~Zhao, C.~Ruan, F.~Luo, and W.~Liang, ``{DeepSeek}-{Coder}-{V2}: {Breaking} the {Barrier} of {Closed}-{Source} {Models} in {Code} {Intelligence},'' June 2024.
\newblock arXiv:2406.11931.

\bibitem{hui_qwen25-coder_2024}
B.~Hui, J.~Yang, Z.~Cui, J.~Yang, D.~Liu, L.~Zhang, T.~Liu, J.~Zhang, B.~Yu, K.~Lu, K.~Dang, Y.~Fan, Y.~Zhang, A.~Yang, R.~Men, F.~Huang, B.~Zheng, Y.~Miao, S.~Quan, Y.~Feng, X.~Ren, X.~Ren, J.~Zhou, and J.~Lin, ``Qwen2.5-{Coder} {Technical} {Report},'' Nov. 2024.
\newblock arXiv:2409.12186.

\bibitem{ai_codestral_2024}
M.~AI, ``Codestral: {Hello}, {World}!,'' May 2024.

\bibitem{bhargava_cpp-ut-bench_2024}
V.~Bhargava, R.~Ghosh, and D.~Dutta, ``{CPP}-{UT}-{Bench}: {Can} {LLMs} {Write} {Complex} {Unit} {Tests} in {C}++?,'' Dec. 2024.
\newblock arXiv:2412.02735.

\bibitem{yi_team_humanevaljl_2024}
{Yi Team}, ``{HumanEval}.jl,'' Feb. 2024.
\newblock original-date: 2024-02-07T11:42:55Z.

\bibitem{gusanidas_rust_2024}
{Gusanidas}, ``Rust {Compiling} {Benchmark},'' Nov. 2024.
\newblock original-date: 2024-11-21T23:10:10Z.

\end{thebibliography}


\appendix

\section*{Acknowledgements}
\emph{We thank the Armed Forces of Ukraine for providing security to complete this work.}
We thank the anonymous reviewers for providing valuable feedback and suggestions that helped to improve this paper.

\end{document}